\documentclass[lettersize,journal]{IEEEtran}
\usepackage{amsmath,amsfonts}
\usepackage{algorithmic}
\usepackage{algorithm}
\usepackage{array}
\usepackage[caption=false,font=normalsize,labelfont=sf,textfont=sf]{subfig}
\usepackage{textcomp}
\usepackage{stfloats}
\usepackage{url}
\usepackage{verbatim}
\usepackage{graphicx}
\usepackage{cite}
\usepackage{orcidlink}
\usepackage{fix-cm}
\usepackage{amsmath,amssymb,amsfonts}
\usepackage{algorithmic}
\usepackage{graphicx,color}
\usepackage{textcomp}
\usepackage{xcolor}
\usepackage{hyperref}
\hypersetup{hidelinks}
\usepackage{algorithm,algorithmic}
\usepackage{booktabs}
\usepackage{multirow}
\usepackage{siunitx}
\usepackage{fancyhdr} 

\begin{document}

\title{SAVe: Self-Supervised Audio-visual Deepfake Detection Exploiting Visual Artifacts and Audio-visual Misalignment}

\author{Sahibzada Adil Shahzad $^{*}$\orcidlink{0009-0000-5591-8423}, Ammarah Hashmi $^{*}$\orcidlink{0000-0002-1973-6902}, Junichi Yamagishi, \orcidlink{0000-0003-2752-3955} ~\IEEEmembership{Senior Member,~IEEE}, Yusuke Yasuda, \orcidlink{0000-0002-2130-747X}~\IEEEmembership{ Member,~IEEE}, Yu Tsao\orcidlink{0000-0001-6956-0418},~\IEEEmembership{Senior Member,~IEEE}, Chia-Wen Lin\orcidlink{0000-0002-9097-2318},~\IEEEmembership{Fellow,~IEEE}, Yan-Tsung Peng\orcidlink{0000-0002-3802-1670},~\IEEEmembership{Senior Member,~IEEE}, and Hsin-Min Wang\orcidlink{0000-0003-3599-5071},~\IEEEmembership{Senior Member,~IEEE}

\thanks{Sahibzada Adil Shahzad is with the Social Networks and Human-Centered Computing Program, Taiwan International Graduate Program, Institute of Information Science, Academia Sinica, and also with the Department of Computer Science, National Chengchi University, Taipei 11529, Taiwan. (e-mail: adilshah275@iis.sinica.edu.tw).

Ammarah Hashmi is with the Social Networks and Human-Centered Computing Program, Taiwan International Graduate Program, Institute of Information Science, Academia Sinica, Taipei 11529, Taiwan, and also with the Institute of Information Systems and Applications, National Tsing Hua University, Hsinchu 300044, Taiwan. (e-mail: hashmiammarah0@gmail.com).

Junichi Yamagishi and Yusuke Yasuda are with the National Institute of Informatics, Tokyo 101-8430, Japan. (e-mail: jyamagis@nii.ac.jp , yasuda@nii.ac.jp).

Chia-Wen Lin is with the Department of Electrical Engineering and the Institute of Communications Engineering, National Tsing Hua University, Hsinchu 300044, Taiwan. (e-mail: cwlin@ee.nthu.edu.tw).

Yan-Tsung Peng is with the Department of Computer Science, National Chengchi University, Taipei 11605, Taiwan (e-mail: ytpeng@cs.nccu.edu.tw).

Yu Tsao is with the Research Center for Information Technology Innovation, Academia Sinica, Taipei 11529, Taiwan. (e-mail:
yu.tsao@citi.sinica.edu.tw).

Hsin-Min Wang is with the Institute of Information Science, Academia Sinica, Taipei 11529, Taiwan. (e-mail: whm@iis.sinica.edu.tw).}
\thanks{\hrulefill \newline$^{*}$ Equal Contribution }}

\maketitle
\thispagestyle{fancy}

\begin{abstract}
Multimodal deepfakes can exhibit subtle visual artifacts and cross-modal inconsistencies, which remain challenging to detect, especially when detectors are trained primarily on curated synthetic forgeries. Such synthetic dependence can introduce dataset and generator bias, limiting scalability and robustness to unseen manipulations. We propose \textit{SAVe}, a self-supervised audio-visual deepfake detection framework that learns entirely on authentic videos. \textit{SAVe} generates on-the-fly, identity-preserving, region-aware self-blended pseudo-manipulations to emulate tampering artifacts, enabling the model to learn complementary visual cues across multiple facial granularities. To capture cross-modal evidence, \textit{SAVe} also models lip-speech synchronization via an audio-visual alignment component that detects temporal misalignment patterns characteristic of audio-visual forgeries. Experiments on FakeAVCeleb and AV-LipSync-TIMIT demonstrate competitive in-domain performance and strong cross-dataset generalization, highlighting self-supervised learning as a scalable paradigm for multimodal deepfake detection.
\end{abstract}


\section{Introduction}
\begin{figure}[!t]
    \centering
    \includegraphics[width=\linewidth, height=0.38\textheight]{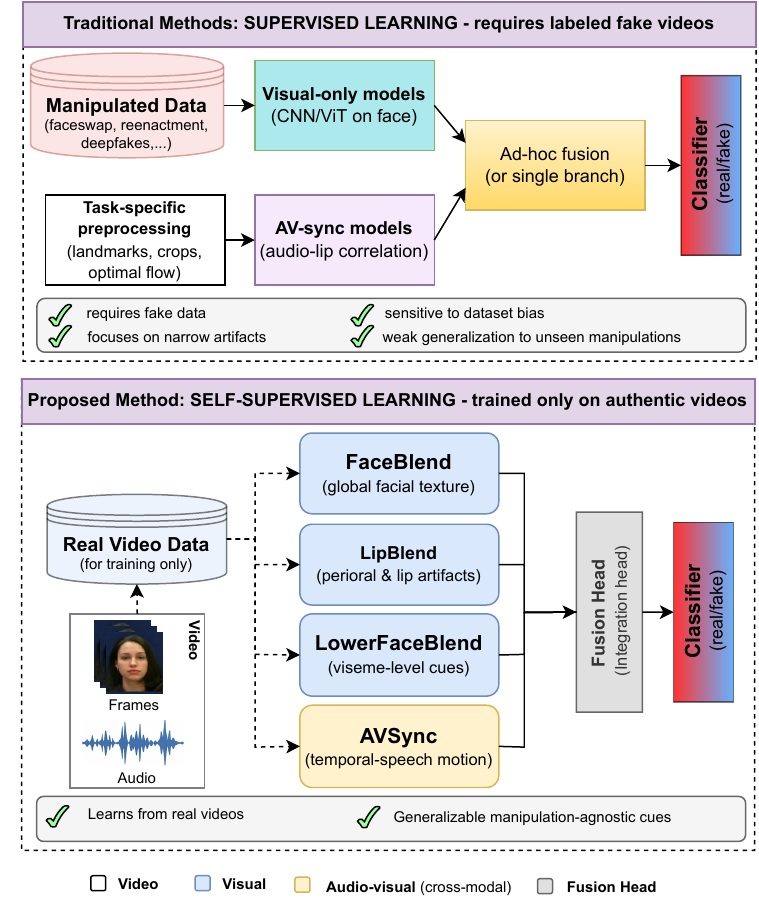}
    \caption{Prior visual/audio-visual deepfake detectors typically rely on manipulated training data and task-specific preprocessing, using either visual-only models or audio-visual synchronization cues with ad-hoc fusion, which can lead to poor generalization under unseen generators. In contrast, the proposed model, SAVe, is trained exclusively on authentic videos by generating on-the-fly, identity-preserving self-blended pseudo-manipulations. SAVe combines \textit{FaceBlend}, \textit{LipBlend}, \textit{LowerFaceBlend}, and \textit{AVSync} features via a Fusion head for robust real/fake prediction.}
    \label{fig:existingwork}
\vspace{-14pt}
\end{figure}

\IEEEPARstart{R}{ecent} progress in audio-visual generative models has increased the realism of talking-face forgeries, motivating detection methods that utilize multimodal evidence rather than relying on uni-modal ones \cite{R1, R2, R56}. Early visual detectors used artifacts such as boundary blending, spatial inconsistencies, and texture anomalies \cite{R2, R3, R4}; however, these cues are increasingly attenuated as the quality of the synthesis improves, shifting attention to speech-related signals and lip-speech correspondence \cite{R5, R6, R7, R8}. Most existing detectors remain fully supervised, as shown in Fig.~\ref{fig:existingwork}, and depend on labeled deepfake datasets, which limits scalability and degrades transfer due to shortcut learning and sensitivity to dataset-specific factors (e.g., compression and preprocessing) rather than manipulation evidence \cite{R9, R10, R11, R12}. This motivates label-efficient, manipulation-agnostic learning, where self-supervised objectives exploit intrinsic temporal and cross-modal consistency without forgery annotations \cite{R13, R14}.

We propose \textit{SAVe}, a self-supervised audio-visual deepfake detection framework trained exclusively on real videos. \textit{SAVe} consists of four complementary modules: \emph{FaceBlend} for generic spatial editing artifacts, \emph{LipBlend} for fine-grained lip-region irregularities, \emph{LowerFaceBlend} for discriminative lower-face cues related to viseme-phoneme coupling, and \emph{AVSync} for temporal synchronization between speech and lip motion. By unifying intra-modal visual consistency and cross-modal audio-visual coherence, \textit{SAVe} 
promotes label-efficient, manipulation-agnostic generalization by discouraging shortcut learning and improving robustness to unseen deepfake generation techniques, while enabling scalable training without labeled forgeries.

Our contributions are as follows.
\begin{itemize}
\item We propose \textit{SAVe}, a self-supervised audio-visual deepfake detection framework trained exclusively on authentic videos, which jointly learns visual manipulation artifacts and audio-visual synchronization inconsistencies without any synthetic deepfake training data.

\item We unify intra-modal and cross-modal consistency objectives within a single framework to improve generalization to diverse and unseen manipulations.

\item We mitigate shortcut learning through manipulation-agnostic consistency cues, enabling label-efficient and cross-dataset robust detection.

\item Extensive experiments on two benchmarks demonstrate strong performance and cross-dataset robustness without forged training data.
\end{itemize} 

\section{RELATED WORK}
Deepfake detection has been studied in visual, audio, and audio-visual settings~\cite{R1, R56}. While supervised detectors achieve strong in-domain performance, their reliance on specific forgery generators and labeling pipelines often induces dataset bias and limits robustness to unseen manipulations. This motivates recent interest in self-supervised and training paradigms that aim to learn intrinsic authenticity cues using authentic data only.

\subsection{Visual deepfake detection}
Visual-only detectors commonly train convolutional neural network (CNN)/Transformer classifiers on manipulated datasets~\cite{R3, R15} and exploit artifacts and inconsistencies (e.g., local texture, facial dynamics, and temporal cues) using explicit sequence modeling~\cite{R16, R17, R18, R19, R20}. Recent Transformer-based designs improve temporal context modeling and attempt to mitigate shortcut cues such as identity leakage, which can otherwise harm cross-dataset generalization~\cite{R21, R22, R23, R24}. Nevertheless, purely visual cues may be unreliable when synthesis is highly photorealistic or when the audio track is manipulated.

\begin{figure*}[!t]
    \centering
    \includegraphics[width=0.9\linewidth,
        height=0.38\textheight]{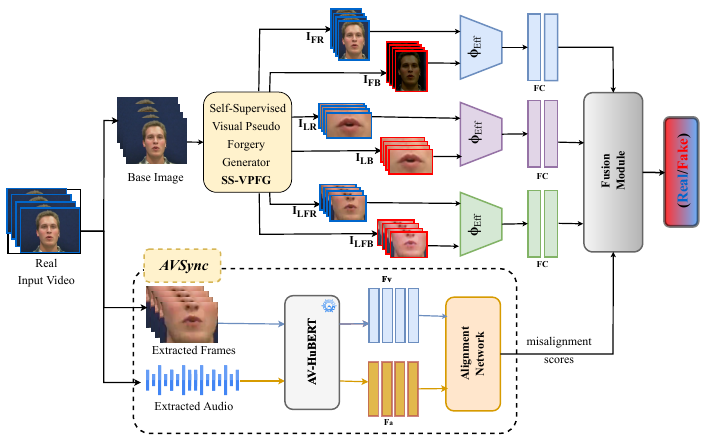}
    \caption{Overview of the proposed \textit{SAVe} framework, a self-supervised learning (SSL) audio-visual deepfake detection method trained solely using authentic data. Given an input video, visual frames and audio are extracted, in the \textit{AVSync} branch, both modalities are encoded by AV-HuBERT to obtain audio ($F_a$) and visual ($F_v$) representations, which are then processed by an alignment network to produce an audio-visual misalignment score that reflects lip-speech synchronization consistency. In parallel, the \textit{SS-VPFG} module synthesizes on-the-fly pseudo-forged visual samples via identity-preserving self-blending over multiple facial regions, enabling region-aware supervision: (face (real $\text{I}_{\text{FR}}$ vs. fake $\text{I}_{\text{FB}}$), lip (real $\text{I}_{\text{LR}}$ vs. fake $\text{I}_{\text{LB}}$), and lower-face (real $\text{I}_{\text{LFR}}$ vs. fake $\text{I}_{\text{LFB}}$)) to provide region-aware supervision. Finally, region-specific visual features and AVSync misalignment cues are aggregated by the Fusion module to output the final prediction (Real/Fake).}
    \label{fig:method}
\vspace{-14pt}
\end{figure*}

\subsection{Audio deepfake detection}
Audio-only spoofing detection targets artifacts introduced by text-to-speech (TTS) or voice conversion (VC) systems and neural vocoders, evolving from hand-crafted spectral features with classical classifiers~\cite{R12} to end-to-end architectures (e.g., RawNet2 and AASIST) and self-supervised speech representations with augmentation for improved robustness~\cite{R26, R27, R28, R29, R31, R32, R33}. Generalization remains challenging under unseen generators, codec distortions, and multilingual conditions, motivating one-class and codec-aware strategies~\cite{R33, R34, R35, R36}. Audio-only approaches also fail when forged video is paired with genuine audio, because they cannot detect cross-modal inconsistencies and visual artifacts.

\subsection{Audio-visual deepfake detection}
Audio-visual detectors exploit speech-face coupling to expose cross-modal mismatches through synchronization constraints, contrastive objectives, and lip-sync modeling~\cite{R7, R37, R38, R39, R40, R57}. Fusion mechanisms such as cross-attention and pretrained audio-visual representations further improve temporal context modeling~\cite{R8, R41, R42, R43}. However, most methods, including LLM-based models that generalize poorly~\cite{R25}, remain supervised on labeled forgeries and may inherit dataset bias from specific manipulation pipelines, limiting transfer to unseen generators and editing strategies.

To reduce dependence on synthetic data, one-class and real-only approaches learn from authentic samples and flag anomalous deviations \cite{R45, R46, R47, R48, R49, R50, R51}. Related self-supervised strategies generate pseudo forgeries from real data using self-blended image techniques \cite{R13, R52}, allowing artifact-aware learning without forged training sets. However, the techniques are designed for broad artifacts across the entire face; consequently, they may miss specific visual artifacts that occur during lip-syncing. Building on this direction, we propose an audio-visual framework that performs region-wise self-blending (full face, lips, and lower face) and incorporates an audio-visual alignment module to capture lip-sync irregularities by leveraging only authentic data. This combination targets both spatial artifacts and cross-modal timing inconsistencies, improving robustness to unseen manipulations without using forged training data.

\section{Proposed Method}
\label{sec:method}


We propose \textit{SAVe}, a self-supervised audio-visual deepfake detector trained exclusively on authentic talking-head videos. \textit{SAVe} learns manipulation-sensitive representations without relying on synthetic deepfake data by combining (i) region-wise self-blended pseudo-manipulations that emulate common compositing artifacts at multiple facial granularities, inspired by~\cite{R13}, and (ii) audio-visual consistency modeling that captures lip-speech synchronization anomalies. Fig.~\ref{fig:method} provides an overview; the Self-Supervised Visual Pseudo-Forgery Generation (SS-VPFG) module generates pseudo-manipulated samples for three visual branches (\textit{FaceBlend} (FB), \textit{LipBlend} (LB), and \textit{LowerFaceBlend} (LFB)) that learn complementary artifact cues at different spatial scales, while an audio-visual branch \textit{AVSync} (based on~\cite{R14}) detects temporal misalignment between speech and lip motion. 
The final authenticity score is obtained via average logit fusion over the four branch predictions (FB, LB, LFB, and AVSync), improving robustness to unseen manipulations.

\subsection{Self-Blended Pseudo Forgeries}
Given a video clip ${V}$ with visual frames $\{x_t\}_{t=1}^{T}$, we create a pseudo forgery from each real frame by blending two stochastically augmented views of the same frame:
\begin{equation}
\begin{aligned}
x_t^{s}=\mathcal{A}_s(x_t), \quad x_t^{t}=\mathcal{A}_t(x_t), \\ 
\tilde{x}_t^{(r)} = x_t^{s}\odot M^{(r)} + x_t^{t}\odot(\textbf{1}-M^{(r)}),
\end{aligned}
\label{eq:selfblend}
\end{equation}
where $x_t^{s}$ and $x_t^{t}$ are, respectively, the \emph{source} and \emph{target} augmented views while \emph{source} and \emph{target} are used for blending,  $A$ is a source-target augmentation, $\odot$ is element-wise multiplication, and $\textbf{1}$ is an all-one mask. $M^{(r)}\!\in[0,1]^{H\times W}$ is a soft blending mask for region $r$, obtained from facial landmarks and randomly deformed/smoothed to mimic imperfect compositing boundaries. We use $r \in \{\text{face}, \text{lip}, \text{lower-face}\}$ to generate training pairs for \textit{FB}, \textit{LB}, and \textit{LFB}, respectively. The original frame $x_t$ is the \emph{base image}, target frame $x_t^{t}$ is treated as \emph{real}, while $\tilde{x}_t^{(r)}$ serves as a \emph{pseudo-manipulated} sample. This self-blending strategy is inspired by~\cite{R13}; however, it is adapted to be \emph{region-aware} by constructing $M^{(r)}$ over different facial granularities. In particular, beyond simulating global facial artifacts commonly associated with face-swapping pipelines, we additionally generate localized perturbations in the mouth area to reflect better fine-grained inconsistencies often introduced by lip-sync manipulations. Overall, the proposed procedure produces controllable artifact patterns (e.g., color/texture discontinuities and boundary inconsistencies) while training entirely on authentic data, thereby reducing synthetic-data bias and improving robustness to unseen manipulations.

\begin{figure*}[!ht]
    \centering
    \includegraphics[width=0.9\linewidth,
        height=0.27\textheight]{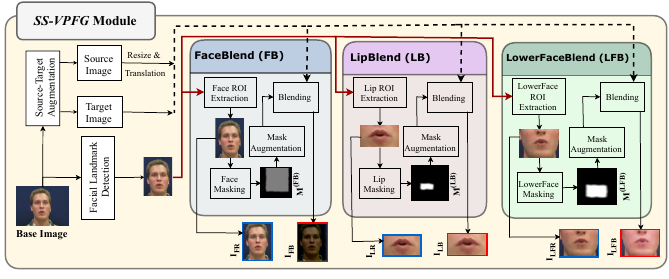}
    \caption{Overview of the proposed Self-Supervised Visual Pseudo-Forgery Generation (SS-VPFG) module. Given a Base Image, the module first performs source-target augmentation and facial landmark detection to produce aligned source and target images. Three region-specific blending pipelines are then applied: \textit{FaceBlend}, \textit{LipBlend}, and \textit{LowerFaceBlend}. Each pipeline extracts the corresponding region of interest (ROI), generates region masks, applies targeted augmentations, and outputs forged variants, namely $\text{I}_{\text{FB}}$, $\text{I}_{\text{LB}}$, and $\text{I}_{\text{LFB}}$, alongside their intermediate augmented results ($\text{I}_{\text{FR}}$, $\text{I}_{\text{LR}}$, and $\text{I}_{\text{LFR}}$). These pseudo forgeries provide diverse self-supervised signals for training visual forgery detectors.}
    \label{fig:ss-vpfg}
\vspace{-14pt}
\end{figure*}

\subsection{Multi-Region Visual Artifact Learning}
To learn manipulation-sensitive visual cues without synthetic forgeries, as shown in Fig.~\ref{fig:ss-vpfg}, we design a \textbf{Self-Supervised Visual Pseudo-Forgery Generator (\textit{SS-VPFG})} coupled with three region-specific visual learners. All branches share the same feature extractor $f_\theta$ (following~\cite{R13}) and differ only in the facial region used to construct the blending mask $M^{(r)}$.
\vspace{-2pt}
\begin{itemize}\setlength\itemsep{1pt}

\item \textbf{\textit{FaceBlend} (FB):}  $M^{(\mathrm{FB})}$ spans the full facial area to emphasize global compositing cues, such as color/illumination discontinuities, texture mismatch, and over-smoothing that often emerge in face synthesis and replacement pipelines.

\item \textbf{\textit{LipBlend} (LB):} $M^{(\mathrm{LB})}$ restricts blending to the lips and adjacent perioral skin, thereby capturing fine-grained artifacts near the vermilion boundary and intra-oral textures (e.g., teeth and tongue), where reenactment and retalking methods often introduce high-frequency distortions and boundary leakage.

\item \textbf{\textit{LowerFaceBlend} (LFB):} $M^{(\mathrm{LFB})}$ covers the mouth-jaw-chin region to capture broader lower-face inconsistencies, including geometric warping and skin-texture mismatch along the jawline, complementing more localized LB cues.

\end{itemize}\vspace{-2pt}
Each branch predicts whether an input is real $x_t$ or self-blended $\tilde{x}_t^{(r)}$. 

\subsection{Audio-Visual Synchronization Consistency}
Visual artifacts alone may miss lip-sync temporal irregularities; thus, we incorporate an auxiliary audio-visual synchronization module, adopted from the originally standalone unsupervised detector proposed in~\cite{R14}, as a temporal cross-modal anomaly detector. We denote this module as \textit{AVSync} to reflect its synchronization-consistency role in \textit{SAVe} framework. We extract frame-rate-aligned audio and visual features $\{a_t\}_{t=1}^{T}$ and $\{v_t\}_{t=1}^{T}$ using a pretrained AV-HuBERT~\cite{R62} encoder, and feed them to a fully connected network $s(a_i,v_j)$, which is a four layers multilayer perceptron (MLP) with layer normalization and ReLU activation function. We enforce that each audio step matches its corresponding visual step within a temporal neighborhood $\mathcal{N}(i)$ using an InfoNCE objective \cite{R58}:
\begin{equation}
\mathcal{L}_{\mathrm{AV}}=-\frac{1}{T}\sum_{i=1}^{T}\log p(v_i|a_i),
\end{equation}
where $p(v_i|a_i)=\frac{\exp(s(a_i,v_i))}{\sum_{j\in\mathcal{N}(i)}\exp(s(a_i,v_j))}$.
In inference, low aggregated alignment scores indicate a synchronized audio-visual modality, while high score shows misalignment. Importantly, \textit{SS-VPFG} remains effective in a \emph{visual-only} setting; \textit{AVSync} provides complementary audio-visual temporal cues when audio is available.

\subsection{Fusion Module}
For each input video, our framework produces four independent prediction scores: three visual predictions from \textit{FB}, \textit{LB}, and \textit{LFB}, and one audio-visual prediction from the \textit{AVSync} module. Each branch outputs a scalar score representing the confidence of the input being fake.

Before fusion, we apply a lightweight, deterministic score normalization to the audio-visual (\textit{AVSync}) branch only, in order to mitigate scale mismatch with the visual branches. Specifically, the \textit{AVSync} scores are normalized using min-max scaling, followed by a logit-sigmoid mapping. This transformation preserves the relative ordering of predictions while constraining the scores to a stable probabilistic range, without introducing additional parameters or supervision.

Final decision-level fusion is performed using a parameter-free average logit fusion strategy. Given the calibrated \textit{AVSync} score and the raw scores from the three visual branches, all scores are first transformed into logit space, averaged across the four branches, and then mapped back to probability space using a sigmoid function. Formally, the fused prediction is computed as
\begin{equation}
p^{(\mathrm{final})}
=
\sigma\!\left(
\frac{1}{4}
\sum_{k \in \{\mathrm{FB}, \mathrm{LB}, \mathrm{LFB}, \mathrm{AV}\}}
\mathrm{logit}\!\left(\tilde{p}^{(k)}\right)
\right),
\end{equation}
where $\tilde{p}^{(k)}$ denotes the calibrated score for the AVSync branch and the raw scores for the visual branches, respectively. The final predicted label is obtained by thresholding $p^{(\mathrm{final})}$.

This late-fusion design enables each branch to capture complementary forgery cues, including regional visual artifacts and audio-visual synchronization inconsistencies, while preserving a simple, robust, and fully parameter-free inference pipeline.

\section{Experiments}
\label{sec:experiments}

This section outlines the experimental protocol, including datasets, evaluation metrics, and training/evaluation settings, and reports both in-domain and cross-dataset results. 

\subsection{Experimental Setup}
We evaluated our proposed method on two widely used audio-visual deepfake benchmarks: FakeAVCeleb~\cite{R53} and AV-LipSync-TIMIT~\cite{R54} (denoted LipSyncTIMIT for brevity). The training process exclusively utilizes real videos from these datasets.

\textbf{FakeAVCeleb} is constructed from real celebrity interview videos (sourced from VoxCeleb2 \cite{R59}) and the corresponding forged variants. It contains 20,000 videos (500 real and 19,500 fake) and features subject selection designed to enhance racial and overall demographic diversity. FakeAVCeleb defines four types of audio-visual pairing: RARV (real audio + real video), FARV (fake audio + real video), RAFV (real audio + fake video), and FAFV (fake audio + fake video). Forged samples are generated using common face manipulation pipelines (e.g., \textit{FaceSwap}, \textit{FSGAN}, and \textit{Wav2Lip}), while voice cloning is produced using SV2TTS; the FAFV subset combines cloned speech with face manipulation/reenactment to yield near lip-synchronized forgeries. Importantly, FARV and RTVC (real-time voice cloning) samples are audio-only manipulations paired with authentic video, which is expected to challenge purely visual detectors.

\textbf{AV-LipSync-TIMIT (LipSyncTIMIT)} focuses on lip-sync forensics under diverse audio conditions. It is derived from 202 real VidTIMIT \cite{R60} videos (43 speakers) and generates lip-sync forgeries using five methods (\textit{Diff2Lip}, \textit{VideoReTalking}, \textit{Wav2Lip}, \textit{Wav2Lip-GAN}, and \textit{IP-LAP}) under three audio settings: (i) source audio, (ii) real external audio (LRS2), and (iii) AI-generated audio (LibriSeVoc). The uncompressed split contains 3,232 videos (202 real and 3,030 fake) and additional compressed versions at \textit{c23} and \textit{c40}, compressed with constant rate factors of 23 and
40, respectively. Considering the uncompressed data together with these compressed counterparts, the dataset comprises 9,090 videos in total.

\subsection{Implementation Details}
All self-supervised visual branches are trained exclusively on authentic videos. Face frames are extracted using Dlib’s frontal face detector and shape predictor, producing region-specific crops corresponding to the full face, lower face, or lip region. Intra-identity, region-wise pseudo forgeries are generated on-the-fly via SS-VPFG, as shown in Fig. \ref{fig:ss-vpfg}, where landmark-guided regions undergo photometric and geometric perturbations before blending. Each visual branch is optimized with a binary cross-entropy loss to discriminate real frames from self-blended counterparts, providing artifact-aware supervision without relying on synthetic deepfakes.

The audio-visual synchronization (\textit{AVSync}) branch follows~\cite{R14}, using AV-HuBERT to extract video and audio features and trained with an InfoNCE objective to maximize correspondence between synchronized pairs. At inference, up to 8 frames per video are uniformly sampled, processed independently, and averaged for video-level scores.  

For evaluation, we report the Area Under the Receiver Operating Characteristic Curve (AUC) and Average Precision (AP). AUC measures threshold-free ranking quality and is relatively robust to class imbalance, while AP complements AUC by emphasizing precision-recall behavior under skewed class distributions, which is common in deepfake benchmarks.



\vspace{-2pt}
\subsection{Evaluations and Analysis}

\subsubsection{In-domain Evaluation of Visual Branches on FakeAVCeleb}
Table~\ref{tab:fakeav_in_domain} reports in-domain results on FakeAVCeleb when training exclusively on authentic videos with on-the-fly, intra-identity self-blended pseudo forgeries. \textit{FaceBlend} achieves the strongest overall performance, reaching near-saturated detection on \textit{Wav2Lip} and \textit{FSGAN} (and their compositional variants) with $\mathrm{AUC}/\mathrm{AP}\approx 1.00/1.00$, while remaining robust on \textit{FaceSwap} ($0.98/0.99$). This suggests that full-face self-blending captures broadly transferable cues that align well with common in-domain generation pipelines.

The region-focused branches (\textit{LowerFaceBlend} and \textit{LipBlend}) also perform strongly, particularly on mouth-centric and compositional forgeries (e.g., \textit{FaceSwap+Wav2Lip}: $1.00/1.00$ and $0.99/1.00$), but exhibit slightly reduced robustness on identity-replacement attacks such as FaceSwap, where artifacts may extend beyond the mouth region.


A key and expected limitation arises on RTVC, where all visual only branches converge to chance level performance. In FakeAVCeleb, RTVC corresponds to audio only manipulation paired with authentic video, such that no visual tampering is present. Consequently, a detector that operates solely on facial frames has no causal visual evidence to exploit. Importantly, the resulting chance level performance does not reflect random behavior, but is a direct consequence of the data composition and labeling. For both real video with fake audio and real video with real audio samples, a visual only model will correctly identify the video content as real, leading to an equal number of correct and incorrect predictions and an AUC of 0.50 by construction. Achieving higher performance in this setting would require the model to exploit spurious correlations or dataset specific shortcuts unrelated to genuine forgery cues. The absence of such gains indicates that the model does not rely on shortcut learning, which is desirable in the context of principled visual forgery detection.

\begin{table}[!t]
\centering
\caption{In-domain performance on FakeAVCeleb for visual self-supervised branches trained \emph{only on real videos} with on-the-fly region-wise self-blending. We report AUC and AP for FaceBlend, LowerFaceBlend, and LipBlend, evaluated per manipulation type.}
\label{tab:fakeav_in_domain}
\scriptsize
\setlength{\tabcolsep}{3.6pt}
\renewcommand{\arraystretch}{1.12}
\sisetup{
  round-mode = places,
  round-precision = 2 
}
\begin{tabular}{l S[table-format=1.3] S[table-format=1.3] S[table-format=1.3] S[table-format=1.3] S[table-format=1.3] S[table-format=1.3]}
\toprule
\multirow{2}{*}{Manipulation} &
\multicolumn{2}{c}{FaceBlend} &
\multicolumn{2}{c}{LowerFaceBlend} &
\multicolumn{2}{c}{LipBlend} \\
\cmidrule(lr){2-3}\cmidrule(lr){4-5}\cmidrule(lr){6-7}
& AUC & AP & AUC & AP & AUC & AP \\
\midrule
Wav2Lip            & 0.9999 & 0.9999 & 0.9895 & 0.9994 & 0.9826 & 0.9981 \\
FaceSwap           & 0.9781 & 0.9859 & 0.9642 & 0.9736 & 0.9718 & 0.9740 \\
FaceSwap+Wav2Lip   & 0.9999 & 0.9999 & 0.9955 & 0.9991 & 0.9924 & 0.9972 \\
FSGAN              & 0.9999 & 0.9999 & 0.9803 & 0.9971 & 0.9736 & 0.9946 \\
FSGAN+Wav2Lip      & 0.9999 & 0.9999 & 0.9957 & 0.9994 & 0.9910 & 0.9972 \\
RTVC               & 0.4634 & 0.4745 & 0.5301 & 0.5328 & 0.5164 & 0.5161 \\
\bottomrule
\end{tabular}
\end{table}

\subsubsection{Cross-dataset Evaluation of Visual Branches: LipSyncTIMIT \texorpdfstring{$\rightarrow$}{→} FakeAVCeleb}
Training on LipSyncTIMIT and evaluating on FakeAVCeleb expose a clear domain shift relative to the in-domain setting, yet the proposed self-supervised branches retain meaningful transferability to several unseen generators and manipulation compositions, as shown in Table~\ref{tab:lipsync_to_fakeav_aucap}. Overall, \textit{LipBlend} demonstrates the strongest cross-dataset robustness on face-swap-centric attacks: it remains consistently high on \textit{FaceSwap} across all training settings (e.g., $0.93/0.95$ when trained on all compressions and $0.92/0.94$ even when trained on heavily compressed \textit{c40}), and it performs particularly well on the composite \textit{FaceSwap+Wav2Lip} pipeline ($0.96/0.99$ with all-compression training). These results support the hypothesis that mouth-region supervision learns cues that are less sensitive to dataset-specific global appearance statistics and more stable under domain shifts.

\begin{table}[!t]
\centering
\caption{Cross-dataset generalization from LipSyncTIMIT to FakeAVCeleb for visual self-supervised branches. Models are trained on LipSyncTIMIT under different compression settings and evaluated on FakeAVCeleb across unseen manipulation types. We report AUC and AP.}
\label{tab:lipsync_to_fakeav_aucap}
\scriptsize
\setlength{\tabcolsep}{3.0pt}
\renewcommand{\arraystretch}{1.08}
\sisetup{
  round-mode = places,
  round-precision = 2 
}
\begin{tabular}{l S[table-format=1.3] S[table-format=1.3] S[table-format=1.3] S[table-format=1.3] S[table-format=1.3] S[table-format=1.3]}
\toprule
\multirow{2}{*}{Method} &
\multicolumn{2}{c}{FaceBlend} &
\multicolumn{2}{c}{LowerFaceBlend} &
\multicolumn{2}{c}{LipBlend} \\
\cmidrule(lr){2-3}\cmidrule(lr){4-5}\cmidrule(lr){6-7}
& AUC & AP & AUC & AP & AUC & AP \\
\midrule
\multicolumn{7}{c}{\textit{Trained on LipSyncTIMIT (All compressions: original, c23, c40)}} \\
\midrule
Wav2Lip           & 0.8266 & 0.9865 & 0.7680 & 0.9825 & 0.8342 & 0.9875 \\
FaceSwap          & 0.6676 & 0.7334 & 0.7978 & 0.8508 & 0.9345 & 0.9536 \\
FaceSwap+Wav2Lip  & 0.8165 & 0.9516 & 0.8560 & 0.9637 & 0.9565 & 0.9898 \\
FSGAN             & 0.8326 & 0.9720 & 0.7854 & 0.9626 & 0.8429 & 0.9755 \\
FSGAN+Wav2Lip     & 0.8823 & 0.9713 & 0.8361 & 0.9713 & 0.9026 & 0.9841 \\
RTVC              & 0.5070 & 0.5100 & 0.5061 & 0.5103 & 0.5082 & 0.5080 \\
\midrule
\multicolumn{7}{c}{\textit{Trained on LipSyncTIMIT (c23)}} \\
\midrule
Wav2Lip           & 0.8714 & 0.9905 & 0.7874 & 0.9829 & 0.8143 & 0.9852 \\
FaceSwap          & 0.7608 & 0.7994 & 0.8593 & 0.8809 & 0.9352 & 0.9526 \\
FaceSwap+Wav2Lip  & 0.8925 & 0.9738 & 0.8943 & 0.9696 & 0.9501 & 0.9878 \\
FSGAN             & 0.8590 & 0.9765 & 0.8148 & 0.9655 & 0.8197 & 0.9702 \\
FSGAN+Wav2Lip     & 0.9200 & 0.9857 & 0.8599 & 0.9730 & 0.8829 & 0.9798 \\
RTVC              & 0.4828 & 0.4996 & 0.5050 & 0.5093 & 0.5076 & 0.5079 \\
\midrule
\multicolumn{7}{c}{\textit{Trained on LipSyncTIMIT (c40)}} \\
\midrule
Wav2Lip           & 0.7113 & 0.9766 & 0.6275 & 0.9671 & 0.6758 & 0.9719 \\
FaceSwap          & 0.6519 & 0.7231 & 0.7603 & 0.8178 & 0.9182 & 0.9433 \\
FaceSwap+Wav2Lip  & 0.6934 & 0.9148 & 0.7432 & 0.9297 & 0.8845 & 0.9710 \\
FSGAN             & 0.7210 & 0.9510 & 0.6832 & 0.9381 & 0.7629 & 0.9611 \\
FSGAN+Wav2Lip     & 0.7688 & 0.9553 & 0.6967 & 0.9390 & 0.7941 & 0.9637 \\
RTVC              & 0.4763 & 0.4842 & 0.5155 & 0.5119 & 0.5128 & 0.5182 \\
\midrule
\multicolumn{7}{c}{\textit{Trained on LipSyncTIMIT (original)}} \\
\midrule
Wav2Lip           & 0.8729 & 0.9907 & 0.7731 & 0.9819 & 0.8135 & 0.9861 \\
FaceSwap          & 0.7597 & 0.8063 & 0.8523 & 0.8822 & 0.9388 & 0.9576 \\
FaceSwap+Wav2Lip  & 0.8979 & 0.9755 & 0.8873 & 0.9704 & 0.9572 & 0.9904 \\
FSGAN             & 0.8594 & 0.9770 & 0.8009 & 0.9650 & 0.8213 & 0.9723 \\
FSGAN+Wav2Lip     & 0.9231 & 0.9866 & 0.8458 & 0.9718 & 0.8856 & 0.9816 \\
RTVC              & 0.4887 & 0.5039 & 0.5126 & 0.5157 & 0.5151 & 0.5194 \\
\bottomrule
\end{tabular}
\end{table}

\begin{table}[!b]
\centering
\caption{In-domain performance of the audio-visual synchronization module 
(\textit{AVSync})
on FakeAVCeleb. We report AUC and AP per manipulation type and over the entire dataset.}
\label{tab:av_sync_fakeav_in_domain}
\scriptsize
\setlength{\tabcolsep}{3.6pt}
\renewcommand{\arraystretch}{1.12}
\sisetup{
  round-mode = places,
  round-precision = 2 
}
\begin{tabular}{l S[table-format=1.3] S[table-format=1.3]}
\toprule
\multirow{2}{*}{Manipulation} & 
\multicolumn{2}{c}{\textit{AVSync}}\\
\cmidrule(lr){2-3}
& AUC & AP \\
\midrule
Wav2Lip            & 0.9623 & 0.9945 \\
FaceSwap           & 0.8630 & 0.8737 \\
FaceSwap+Wav2Lip   & 0.9558 & 0.9800 \\
FSGAN              & 0.8863 & 0.9757 \\
FSGAN+Wav2Lip      & 0.9477 & 0.9834 \\
RTVC               & 0.9936 & 0.9884 \\
\midrule
Entire Dataset     & 0.9414 & 0.9974 \\
\bottomrule
\end{tabular}
\end{table}

\textit{FaceBlend} is comparatively stronger on lip-sync-driven manipulations when the training data are cleaner (e.g., \textit{Wav2Lip}: $0.87/0.99$ for \textit{c23} and $0.87/0.99$ for \textit{original}) but degrades more noticeably on pure identity-replacement \textit{FaceSwap} ($0.67/0.73$ with all compressions). \textit{LowerFaceBlend} generally lies between these extremes, improving over \textit{FaceBlend} on \textit{FaceSwap} (up to $0.86/0.88$ for \textit{c23}) while remaining competitive on mixed attacks.

Compression mismatch further affects generalization: training on \textit{c40} consistently lowers AUC for all visual branches (e.g., \textit{Wav2Lip} drops to $0.71$, $0.63$, and $0.68$ for \textit{FaceBlend}, \textit{LowerFaceBlend}, and \textit{LipBlend}, respectively), indicating that heavy compression suppresses subtle spatial artifacts and weakens ranking-based discrimination. Finally, \textit{RTVC} remains near chance for all visual branches, which is expected because \textit{RTVC} is an audio-only manipulation paired with an authentic video stream. 
Collectively, these findings indicate that the visual modality alone, whether relying on global facial cues or localized mouth and lip features, is insufficient to reliably capture synthetic artifacts; this limitation motivates the incorporation of explicit audio-visual synchronization and consistency modeling within the complete framework.

\subsubsection{In-domain Evaluation of \textit{AVSync} on FakeAVCeleb}

Table~\ref{tab:av_sync_fakeav_in_domain} summarizes the in-domain performance of \textit{AVSync}, which detects forgeries by modeling temporal consistency between speech audio and lip motion. \textit{AVSync} achieves strong dataset-level separability $\mathrm{AUC}=0.94$ and $\mathrm{AP}=1.00$, indicating that synchronization cues alone are highly discriminative.
It performs best on lip-sync-related manipulations, including \textit{Wav2Lip}($0.96/0.99$) and compositional attacks (\textit{FaceSwap+Wav2Lip}: $0.96/0.98$; \textit{FSGAN+Wav2Lip}: $0.95/0.98$), while identity-driven manipulations that largely preserve alignments are more challenging (e.g.,  \textit{FaceSwap}: $0.86/0.87$; \textit{FSGAN}: $0.89/0.98$). Notably, \textit{AVSync} excels on the audio-only \textit{RTVC} attack ($0.99/0.99$), making it a complementary auxiliary branch to visual artifact-based detectors.

\begin{table}[!b]
\centering
\caption{Cross-dataset generalization of \textit{AVSync} from LipSyncTIMIT to FakeAVCeleb. \textit{AVSync} is trained on LipSyncTIMIT under different compression settings and evaluated on FakeAVCeleb. We report AUC and AP.}
\label{tab:av_sync_lipsync_to_fakeav_aucap}
\scriptsize
\setlength{\tabcolsep}{3.0pt}
\renewcommand{\arraystretch}{1.08}
\sisetup{
  round-mode = places,
  round-precision = 2 
}
\begin{tabular}{l S[table-format=1.3] S[table-format=1.3]}
\toprule
\multirow{2}{*}{Manipulation} & 
\multicolumn{2}{c}{\textit{AVSync}} \\
\cmidrule(lr){2-3}
& AUC & AP \\
\midrule
\multicolumn{3}{c}{\textit{Trained on LipSyncTIMIT (original)}} \\
\midrule
Wav2Lip           & 0.8920 & 0.9893 \\
FaceSwap          & 0.6874 & 0.7644 \\
FaceSwap+Wav2Lip  & 0.8791 & 0.9585 \\
FSGAN             & 0.7165 & 0.9475 \\
FSGAN+Wav2Lip     & 0.8681 & 0.9677 \\
RTVC              & 0.9398 & 0.9147 \\
\midrule
Entire Dataset    & 0.8462 & 0.9939 \\
\midrule
\multicolumn{3}{c}{\textit{Trained on LipSyncTIMIT (c23)}} \\
\midrule
Wav2Lip           & 0.8924 & 0.9892 \\
FaceSwap          & 0.6786 & 0.7579 \\
FaceSwap+Wav2Lip  & 0.8824 & 0.9598 \\
FSGAN             & 0.7053 & 0.9450 \\
FSGAN+Wav2Lip     & 0.8699 & 0.9682 \\
RTVC              & 0.9436 & 0.9195 \\
\midrule
Entire Dataset    & 0.8447 & 0.9939 \\
\midrule
\multicolumn{3}{c}{\textit{Trained on LipSyncTIMIT (c40)}} \\
\midrule
Wav2Lip           & 0.7100 & 0.9669 \\
FaceSwap          & 0.5492 & 0.6581 \\
FaceSwap+Wav2Lip  & 0.6702 & 0.8783 \\
FSGAN             & 0.5304 & 0.9056 \\
FSGAN+Wav2Lip     & 0.6521 & 0.9059 \\
RTVC              & 0.8386 & 0.7743 \\
\midrule
Entire Dataset    & 0.6575 & 0.9838 \\
\bottomrule
\end{tabular}
\end{table}

\subsubsection{Cross-dataset Evaluation of \textit{AVSync}: LipSyncTIMIT \texorpdfstring{$\rightarrow$}{→} FakeAVCeleb}

Table~\ref{tab:av_sync_lipsync_to_fakeav_aucap} evaluates cross-dataset generalization when \textit{AVSync} is trained on LipSyncTIMIT and tested on FakeAVCeleb. Under clean-to-moderate training conditions (\textit{original} and \textit{c23}), \textit{AVSync} transfers well at the dataset level ($\mathrm{AUC}\approx 0.85$, $\mathrm{AP}=0.99$), suggesting that the learned synchronization prior generalizes across speakers and capture conditions. Generalization is strongest for attacks that inherently break audio-visual correspondence, particularly \textit{RTVC} ($0.94/0.91$ for \textit{original}, $0.94/0.92$ for \textit{c23}), and remains robust on lip-sync/compositional manipulations such as \textit{Wav2Lip} ($\approx 0.89/0.99$) and \textit{FaceSwap+Wav2Lip} ($\approx 0.88/0.96$). Conversely, identity-centric manipulations that may preserve plausible alignment (\textit{FaceSwap}, \textit{FSGAN}) yield substantially lower AUCs ($\approx 0.68$ and $\approx 0.71$), consistent with the limitation of synchronization-only detection. Training under heavy compressions (\textit{c40}) degrades transfer across categories (entire dataset $\mathrm{AUC}=0.66$), indicating that severe compression corrupts cues needed for reliable audio-visual alignment.



\begin{table*}[!t]
\centering
\caption{Comparison with baselines (AUC). Supervised methods (sup.) are trained on labeled real and fake data. Self-supervised methods (unsup.) are trained solely on real videos. AVSync is a component branch of \textit{SAVe}, and Proposed (ours) denotes average-logit fusion of \textit{FB/LB/LFB/AVSync}. 
}
\label{tab:compare_supervised_baselines}
\scriptsize
\setlength{\tabcolsep}{4.6pt}
\renewcommand{\arraystretch}{1.12}
\sisetup{
  round-mode = places,
  round-precision = 2 
}
\begin{tabular}{lcc S[table-format=1.2] S[table-format=1.2] S[table-format=1.2] S[table-format=1.2]}
\toprule
& & & & \multicolumn{3}{c}{LipSyncTIMIT} \\
\cmidrule(lr){5-7}
Model & Train Type & Modality & \multicolumn{1}{c}{FakeAVCeleb-LS} & \multicolumn{1}{c}{Original} & \multicolumn{1}{c}{c23} & \multicolumn{1}{c}{c40} \\
\midrule
Xception~\cite{R15}       & sup.   & V & 0.90   & 0.90   & 0.85   & 0.56 \\
LSDA~\cite{R61}           & sup.   & V &  \text{-}  & 0.91   & 0.84   & 0.60 \\
FTCN~\cite{R20}           & sup.   & V & 0.97   & 0.84   & 0.84   & 0.59 \\
LIPINC-V2~\cite{R54}      & sup.   & V & 0.99   & 0.97   & 0.96   & 0.76 \\
\midrule
FaceBlend      & unsup. & V & 0.9999 & 0.8508 & 0.8047 & 0.4856 \\
LowerFaceBlend & unsup. & V & 0.9895 & 0.8485 & 0.8103 & 0.4716 \\
LipBlend       & unsup. & V & 0.9826 & 0.8142 & 0.7609 & 0.4816 \\
\midrule
AVAD~\cite{R47}           & unsup. & AV & 0.94   & 0.96   & 0.94   & 0.81 \\
AVSync                   & unsup.  & AV & 0.96 & 0.99 & 0.99 & 0.98 \\
\textbf{Proposed (ours)} & unsup.  & AV & 0.9945 & 0.9633 & 0.9688 & 0.7680 \\
\bottomrule
\end{tabular}
\end{table*}

\subsubsection{Comparison with Baseline Methods}

Table \ref{tab:compare_supervised_baselines} presents the AUC performance of the proposed method in comparison with representative supervised and self-supervised baselines on FakeAVCeleb-LS and LipSyncTIMIT under different compression settings. The supervised baselines include Xception~\cite{R15}, LSDA~\cite{R61}, FTCN~\cite{R20}, and LIPINC-V2~\cite{R54}, all of which are trained using labeled real and fake data and rely exclusively on visual information. In contrast, the proposed approach follows a self-supervised learning paradigm and does not require fake samples during training. All results on FakeAVCeleb-LS are obtained using models trained only on the real videos from FakeAVCeleb, whereas results on LipSyncTIMIT correspond to models trained on the original (raw) real videos from the AV-LipSyncTIMIT dataset.

Among the supervised methods, LIPINC-V2 achieves the strongest overall performance, particularly on the compressed LipSyncTIMIT variants, demonstrating the effectiveness of lip-focused visual features when explicit supervision is available. Nevertheless, a noticeable performance degradation is observed under strong compression, especially for the c40 setting, indicating limited robustness to severe codec artifacts.

The self-supervised visual baselines, namely \textit{FaceBlend}, \textit{LowerFaceBlend}, and \textit{LipBlend}, achieve very high performance on FakeAVCeleb-LS, with an AUC value of 0.99. However, their performance drops substantially on LipSyncTIMIT, particularly under compression. This behavior suggests that purely visual self-supervised cues are prone to dataset-specific biases and are less robust to domain shifts introduced by compression.

Self-supervised audio-visual methods exhibit improved robustness across datasets. AVAD~\cite{R47} consistently outperforms visual-only self-supervised baselines on all LipSyncTIMIT variants, highlighting the advantage of exploiting cross-modal consistency. The \textit{AVSync} branch achieves near-perfect performance on LipSyncTIMIT across all compression levels, but its generalization to FakeAVCeleb-LS is comparatively weaker, indicating sensitivity to dataset characteristics.

The proposed method integrates \textit{FaceBlend}, \textit{LowerFaceBlend}, \textit{LipBlend}, and \textit{AVSync} through average logit fusion, achieving a strong balance between generalization and robustness. It obtains an AUC of 0.99 on FakeAVCeleb-LS while maintaining high performance on LipSyncTIMIT across the original, c23, and c40 settings. These results demonstrate that combining complementary visual forgery cues with audio-visual synchronization information leads to more stable and generalizable detection performance than any individual branch alone, without relying on supervised fake data. Overall, the results confirm that the proposed self-supervised audiovisual fusion framework effectively narrows the performance gap with fully supervised methods while offering improved robustness across datasets and compression conditions.



\section{CONCLUSION}
In this paper, we introduce \textbf{\textit{SAVe}}, a fully self-supervised framework for audio-visual deepfake detection that learns authenticity-aware representations from real talking-head videos, without labeled synthetic data. \textit{SAVe} comprises four complementary branches, \textit{FaceBlend}, \textit{LipBlend}, 
\textit{LowerFaceBlend},
and \textit{AVSync}, which jointly capture spatial, articulatory, and temporal coherence between facial motion and speech. By generating pseudo forgeries via region-specific self-blending, the model is systematically exposed to universal forgery cues (e.g., color/frequency shifts, boundary artifacts, and speech-lip misalignments) while preserving real background context. Across benchmarks, \textit{SAVe} achieves strong in-distribution accuracy and markedly better generalization to unseen manipulations, outperforming state-of-the-art visual-only and audio-visual baselines. These results highlight \textit{SAVe} as a scalable, data-efficient, and practical defense against the growing threat of deepfake videos.

\section{ACKNOWLEDGMENT}
This study is supported by JST AIP Acceleration Research (JPMJCR24U3), Japan.

\bibliographystyle{IEEEtran}   
\bibliography{refs} 

@article{R1,
  title={A survey on multimedia-enabled deepfake detection: state-of-the-art tools and techniques, emerging trends, current challenges \& limitations, and future directions},
  author={Khan, Abdullah Ayub and Laghari, Asif Ali and Inam, Syed Azeem and Ullah, Sajid and Shahzad, Muhammad and Syed, Darakhshan},
  journal={Discover Computing},
  volume={28},
  number={1},
  pages={48},
  year={2025}
}

@inproceedings{R2,
  title={Exploiting visual artifacts to expose deepfakes and face manipulations},
  author={Matern, Falko and Riess, Christian and Stamminger, Marc},
  booktitle={Proc. of the IEEE Winter Applications of Computer Vision Workshops},
  pages={83--92},
  year={2019}
}

@inproceedings{R3,
  title={{Mesonet}: a compact facial video forgery detection network},
  author={Afchar, Darius and Nozick, Vincent and Yamagishi, Junichi and Echizen, Isao},
  booktitle={Proc. of the IEEE International Workshop on Information Forensics and Security},
  pages={1--7},
  year={2018}
}

@inproceedings{R4,
  title={Exposing DeepFake Videos By Detecting Face Warping Artifacts},
  author={Yuezun Li and Siwei Lyu},
  booktitle={Proc. of the CVPR Workshops},
  pages={46--52},
  year={2018}
}

@inproceedings{R5,
  title     = {{Detection of Cross-Dataset Fake Audio Based on Prosodic and Pronunciation Features}},
  author    = {Chenglong Wang and Jiangyan Yi and Jianhua Tao and Chu Yuan Zhang and Shuai Zhang and Xun Chen},
  year      = {2023},
  booktitle = {{Interspeech 2023}},
  pages     = {3844--3848},
}

@inproceedings{R6,
  title={Phoneme-Level Feature Discrepancies: A Key to Detecting Sophisticated Speech Deepfakes},
  author={Zhang, Kuiyuan and Hua, Zhongyun and Lan, Rushi and Zhang, Yushu and Guo, Yifang},
  booktitle={Proc. of the AAAI Conference on Artificial Intelligence},
  volume={39},
  number={1},
  pages={1066--1074},
  year={2025}
}

@article{R7,
  title={{AV-Lip-Sync+}: Leveraging AV-HuBERT to Exploit Multimodal Inconsistency for Deepfake Detection of Frontal Face Videos}, 
  author={Shahzad, Sahibzada Adil and Hashmi, Ammarah and Peng, Yan-Tsung and Tsao, Yu and Wang, Hsin-Min},
  journal={IEEE Transactions on Human-Machine Systems}, 
  year={2025},
  volume={55},
  number={6},
  pages={973--982}
}

@article{R8,
  title={{AVTENet}: A Human-Cognition-Inspired Audio-Visual Transformer-Based Ensemble Network for Video Deepfake Detection},
  author={Hashmi, Ammarah and Shahzad, Sahibzada Adil and Lin, Chia-Wen and Tsao, Yu and Wang, Hsin-Min},
  journal={IEEE Transactions on Cognitive and Developmental Systems},
  year={2025},
  volume={17},
  number={6},
  pages={1360--1376}
}

@article{R9,
  title={{DeepFake video detection}: Insights into model generalisation-A Systematic review},
  author={Ramanaharan, Ramcharan and Guruge, Deepani B and Agbinya, Johnson I},
  journal={Data and Information Management},
  pages={100099},
  year={2025}
}

@article{R10,
  title={On the Effectiveness of Dataset Alignment for Fake Image Detection},
  author={Rajan, Anirudh Sundara and Ojha, Utkarsh and Schloesser, Jedidiah and Lee, Yong Jae},
  journal={arXiv e-prints},
  pages={arXiv--2410},
  year={2024}
}

@inproceedings{R11,
  title={Is synthetic voice detection research going into the right direction?},
  author={Borz{\`\i}, Stefano and Giudice, Oliver and Stanco, Filippo and Allegra, Dario},
  booktitle={Proc. of the IEEE/CVF Conference on Computer Vision and Pattern Recognition},
  pages={71--80},
  year={2022}
}

@article{R12,
  title={{ASVspoof 2019}: A large-scale public database of synthesized, converted and replayed speech},
  author={Wang, Xin and others},
  journal={Computer Speech \& Language},
  volume={64},
  pages={101114},
  year={2020}
}

@inproceedings{R13,
  title={Detecting deepfakes with self-blended images},
  author={Shiohara, Kaede and Yamasaki, Toshihiko},
  booktitle={Proc. of the IEEE/CVF conference on computer vision and pattern recognition},
  pages={18720--18729},
  year={2022}
}

@inproceedings{R14,
  title={Circumventing shortcuts in audio-visual deepfake detection datasets with unsupervised learning},
  author={Smeu, Stefan and Boldisor, Dragos-Alexandru and Oneata, Dan and Oneata, Elisabeta},
  booktitle={Proc. of the Computer Vision and Pattern Recognition Conference},
  pages={18815--18825},
  year={2025}
}

@inproceedings{R15,
  title={{Faceforensics++}: Learning to detect manipulated facial images},
  author={Rossler, Andreas and others},
  booktitle={Proc. of the IEEE/CVF international conference on computer vision},
  pages={1--11},
  year={2019}
}

@inproceedings{R16,
  title={{Lips don't lie}: A generalisable and robust approach to face forgery detection},
  author={Haliassos, Alexandros and Vougioukas, Konstantinos and Petridis, Stavros and Pantic, Maja},
  booktitle={Proc. of the IEEE/CVF conference on computer vision and pattern recognition},
  pages={5039--5049},
  year={2021}
}

@inproceedings{R17,
  title={Protecting world leaders against deep fakes.},
  author={Agarwal, Shruti and Farid, Hany and Gu, Yuming and He, Mingming and Nagano, Koki and Li, Hao},
  booktitle={Proc. of the CVPR Workshops},
  volume={1},
  number={38},
  pages={38--45},
  year={2019}
}

@article{R18,
  title={{Deepvision}: Deepfakes detection using human eye blinking pattern},
  author={Jung, Tackhyun and Kim, Sangwon and Kim, Keecheon},
  journal={IEEE Access},
  volume={8},
  pages={83144--83154},
  year={2020}
}

@article{R19,
  title={Deepfake detection using spatiotemporal transformer},
  author={Kaddar, Bachir and others},
  journal={ACM Transactions on Multimedia Computing, Communications and Applications},
  volume={20},
  number={11},
  pages={1--21},
  year={2024}
}

@inproceedings{R20,
  title={Exploring temporal coherence for more general video face forgery detection},
  author={Zheng, Yinglin and Bao, Jianmin and Chen, Dong and Zeng, Ming and Wen, Fang},
  booktitle={Proc. of the IEEE/CVF international conference on computer vision},
  pages={15044--15054},
  year={2021}
}

@article{R21,
  title={Deepfake video detection using convolutional vision transformer},
  author={Wodajo, Deressa and Atnafu, Solomon},
  journal={arXiv preprint arXiv:2102.11126},
  year={2021}
}

@inproceedings{R22,
  title={Improved Deepfake Video Detection Using Convolutional Vision Transformer},
  author={Deressa, Deressa Wodajo and others},
  booktitle={IEEE Gaming, Entertainment, and Media Conference},
  pages={1--6},
  year={2024}
}

@inproceedings{R23,
  title={Domain generalization for face forgery detection by style transfer},
  author={Kim, Taehoon and Choi, Jongwook and Cho, Hyunjin and Lim, HyoungJun and Choi, Jongwon},
  booktitle={Proc. of the IEEE International Conference on Consumer Electronics},
  pages={1--5},
  year={2024}
}

@inproceedings{R24,
  title={{Implicit identity leakage}: The stumbling block to improving deepfake detection generalization},
  author={Dong, Shichao and others},
  booktitle={Proc. of the IEEE/CVF conference on computer vision and pattern recognition},
  pages={3994--4004},
  year={2023}
}

@article{R25,
  title={{How good is chatgpt at audiovisual deepfake detection}: A comparative study of chatgpt, ai models and human perception},
  author={Shahzad, Sahibzada Adil and Hashmi, Ammarah and Peng, Yan-Tsung and Tsao, Yu and Wang, Hsin-Min and others},
  journal={APSIPA Transactions on Signal and Information Processing},
  volume={14},
  number={1},
  year={2025}
}

@article{R26,
  title={STC antispoofing systems for the ASVspoof2019 challenge},
  author={Lavrentyeva, Galina and others},
  pages={1033--1037},
  journal={Interspeech},
  year={2019}
}

@inproceedings{R27,
  title={End-to-end anti-spoofing with rawnet2},
  author={Tak, Hemlata and others},
  booktitle={Proc. of the IEEE International Conference on Acoustics, Speech and Signal Processing},
  pages={6369--6373},
  year={2021}
}

@inproceedings{R28,
  title={{Aasist}: Audio anti-spoofing using integrated spectro-temporal graph attention networks},
  author={Jung, Jee-weon and others},
  booktitle={Proc. of the IEEE international conference on acoustics, speech and signal processing},
  pages={6367--6371},
  year={2022}
}

@article{R29,
  title={Automatic speaker verification spoofing and deepfake detection using wav2vec 2.0 and data augmentation},
  author={Tak, Hemlata and others},
  journal={The Speaker and Language Recognition Workshop},
  pages={112--119},
  year={2022}
}

@inproceedings{R31,
  title={Relative phase information for detecting human speech and spoofed speech.},
  author={Wang, Longbiao and Yoshida, Yohei and Kawakami, Yuta and Nakagawa, Seiichi},
  booktitle={Interspeech},
  pages={2092--2096},
  year={2015}
}

@article{R32,
  title={Synthetic speech detection using fundamental frequency variation and spectral features},
  author={Pal, Monisankha and Paul, Dipjyoti and Saha, Goutam},
  journal={Computer Speech \& Language},
  volume={48},
  pages={31--50},
  year={2018}
}

@article{R33,
  title={{ASVspoof 2021}: accelerating progress in spoofed and deepfake speech detection},
  author={Yamagishi, Junichi and others},
  journal={Automatic Speaker Verification and Spoofing Countermeasures Challenge},
  pages={47--54},
  year={2021}
}

@article{R34,
  title={A study on data augmentation in voice anti-spoofing},
  author={Cohen, Ariel and Rimon, Inbal and Aflalo, Eran and Permuter, Haim H},
  journal={Speech Communication},
  volume={141},
  pages={56--67},
  year={2022}
}

@article{R35,
  title={One-class learning towards synthetic voice spoofing detection},
  author={Zhang, You and Jiang, Fei and Duan, Zhiyao},
  journal={IEEE Signal Processing Letters},
  volume={28},
  pages={937--941},
  year={2021}
}

@article{R36,
  title={Measuring the Robustness of Audio Deepfake Detectors},
  author={Li, Xiang and Chen, Pin-Yu and Wei, Wenqi},
  journal={arXiv preprint arXiv:2503.17577},
  year={2025}
}

@inproceedings{R37,
  title={{Emotions don't lie}: An audio-visual deepfake detection method using affective cues},
  author={Mittal, Trisha and others},
  booktitle={Proc. of the ACM international conference on multimedia},
  pages={2823--2832},
  year={2020}
}

@inproceedings{R38,
  title={Not made for each other-audio-visual dissonance-based deepfake detection and localization},
  author={Chugh, Komal and Gupta, Parul and Dhall, Abhinav and Subramanian, Ramanathan},
  booktitle={Proc. of the 28th ACM international conference on multimedia},
  pages={439--447},
  year={2020}
}

@inproceedings{R39,
  title={Detecting deep-fake videos from phoneme-viseme mismatches},
  author={Agarwal, Shruti and Farid, Hany and Fried, Ohad and Agrawala, Maneesh},
  booktitle={Proc. of the IEEE/CVF conference on computer vision and pattern recognition workshops},
  pages={660--661},
  year={2020}
}

@article{R40,
  title={Voice-face homogeneity tells deepfake},
  author={Cheng, Harry and others},
  journal={ACM Transactions on Multimedia Computing, Communications and Applications},
  volume={20},
  number={3},
  pages={1--22},
  year={2023}
}

@inproceedings{R41,
  title={Joint audio-visual deepfake detection},
  author={Zhou, Yipin and Lim, Ser-Nam},
  booktitle={Proc. of the IEEE/CVF international conference on computer vision},
  pages={14800--14809},
  year={2021}
}

@inproceedings{R42,
  title={Multimodal forgery detection using ensemble learning},
  author={Hashmi, Ammarah and Shahzad, Sahibzada Adil and others},
  booktitle={Proc. of the Asia-Pacific Signal and Information Processing Association Annual Summit and Conference},
  pages={1524--1532},
  year={2022}
}

@article{R43,
  title={{AVFakeNet}: A unified end-to-end Dense Swin Transformer deep learning model for audio--visual deepfakes detection},
  author={Ilyas, Hafsa and Javed, Ali and Malik, Khalid Mahmood},
  journal={Applied Soft Computing},
  volume={136},
  pages={110124},
  year={2023}
}

@inproceedings{R45,
  title={{Lost in translation}: Lip-sync deepfake detection from audio-video mismatch},
  author={Bohacek, Matyas and Farid, Hany},
  booktitle={Proc. of the IEEE/CVF Conference on Computer Vision and Pattern Recognition},
  pages={4315--4323},
  year={2024}
}

@inproceedings{R46,
  title={Zero-Shot Fake Video Detection by Audio-Visual Consistency},
  author={Li, Xiaolou and Liu, Zehua and Chen, Chen and Li, Lantian and Guo, Li and Wang, Dong},
  booktitle={Proc. of Interspeech},
  pages={2935--2939},
  year={2024}
}

@inproceedings{R47,
  title={Self-supervised video forensics by audio-visual anomaly detection},
  author={Feng, Chao and Chen, Ziyang and Owens, Andrew},
  booktitle={Proc. of the IEEE/CVF conference on computer vision and pattern recognition},
  pages={10491--10503},
  year={2023}
}

@inproceedings{R48,
  title={Training-free deepfake voice recognition by leveraging large-scale pre-trained models},
  author={Pianese, Alessandro and Cozzolino, Davide and Poggi, Giovanni and Verdoliva, Luisa},
  booktitle={Proc. of the ACM Workshop on Information Hiding and Multimedia Security},
  pages={289--294},
  year={2024}
}

@article{R49,
  title={Detecting deepfakes without seeing any},
  author={Reiss, Tal and Cavia, Bar and Hoshen, Yedid},
  journal={arXiv preprint arXiv:2311.01458},
  year={2023}
}

@inproceedings{R50,
  title={Zero-shot detection of ai-generated images},
  author={Cozzolino, Davide and Poggi, Giovanni and Nie{\ss}ner, Matthias and Verdoliva, Luisa},
  booktitle={European Conference on Computer Vision},
  pages={54--72},
  year={2024}
}

@inproceedings{R51,
  title={{Aeroblade}: Training-free detection of latent diffusion images using autoencoder reconstruction error},
  author={Ricker, Jonas and Lukovnikov, Denis and Fischer, Asja},
  booktitle={Proc. of the IEEE/CVF Conference on Computer Vision and Pattern Recognition},
  pages={9130--9140},
  year={2024}
}

@article{R52,
  title={{FSBI}: Deepfake detection with frequency enhanced self-blended images},
  author={Hasanaath, Ahmed Abul and Luqman, Hamzah and Katib, Raed and Anwar, Saeed},
  journal={Image and Vision Computing},
  volume={154},
  pages={105418},
  year={2025}
}

@inproceedings{R53,
  title={{FakeAVCeleb}: A Novel Audio-Video Multimodal Deepfake Dataset},
  author={Khalid, Hasam and Tariq, Shahroz and Kim, Minha and Woo, Simon S},
  booktitle={Proc. of the NeurIPS Datasets and Benchmarks},
pages={1--14},
  year={2021}
}

@article{R54,
  title={Detecting Lip-Syncing Deepfakes: Vision Temporal Transformer for Analyzing Mouth Inconsistencies},
  author={Datta, Soumyya Kanti and Jia, Shan and Lyu, Siwei},
  journal={arXiv preprint arXiv:2504.01470},
  year={2025}
}

@article{R56,
  title={{Understanding Audiovisual Deepfake Detection}: Techniques, Challenges, Human Factors and Perceptual Insights},
  author={Hashmi, Ammarah and Shahzad, Sahibzada Adil and Lin, Chia-Wen and Tsao, Yu and Wang, Hsin-Min},
  journal={arXiv preprint arXiv:2411.07650},
  year={2024}
}

@inproceedings{R57,
  title={{Lip sync matters}: A novel multimodal forgery detector},
  author={Shahzad, Sahibzada Adil and Hashmi, Ammarah and Khan, Sarwar and Peng, Yan-Tsung and Tsao, Yu and Wang, Hsin-Min},
  booktitle={Asia-Pacific Signal and Information Processing Association Annual Summit and Conference},
  pages={1885--1892},
  year={2022}
}

@article{R58,
  title={Representation learning with contrastive predictive coding},
  author={Oord, Aaron van den and Li, Yazhe and Vinyals, Oriol},
  journal={arXiv preprint arXiv:1807.03748},
  year={2018}
}

@article{R59,
  title={{VoxCeleb2}: Deep speaker recognition},
  author={Chung, J and Nagrani, A and Zisserman, A},
  journal={Interspeech},
  pages={1086--1090},
  year={2018}
}

@article{R60,
  title={The vidtimit database},
  author={Sanderson, Conrad},
  journal={Idiap Communication},
  pages={02--06},
  year={2002}
}

@inproceedings{R61,
  title={Transcending forgery specificity with latent space augmentation for generalizable deepfake detection},
  author={Yan, Zhiyuan and Luo, Yuhao and Lyu, Siwei and Liu, Qingshan and Wu, Baoyuan},
  booktitle={Proc. of the IEEE/CVF Conference on Computer Vision and Pattern Recognition},
  pages={8984--8994},
  year={2024}
}

@inproceedings{R62,
  title={Learning Audio-Visual Speech Representation by Masked Multimodal Cluster Prediction},
  author={Shi, Bowen and Hsu, Wei-Ning and Lakhotia, Kushal and Mohamed, Abdelrahman},
  booktitle={Proc. of the ICLR},
  year={2021}
}
\end{document}